\documentclass[lettersize,journal]{IEEEtran}
\usepackage{amsmath,amsfonts}
\usepackage{algorithmic}
\usepackage{algorithm}
\usepackage{array}
\usepackage[caption=false,font=normalsize,labelfont=sf,textfont=sf]{subfig}
\usepackage{textcomp}
\usepackage{stfloats}
\usepackage{url}
\usepackage{verbatim}
\usepackage{graphicx}
\usepackage{cite}
\usepackage{multirow}

\begin{document}

\title{ChildDiffusion: Unlocking the Potential of Generative AI and Controllable Augmentations for Child Facial Data using Stable Diffusion and Large Language Models}

\author{Muhammad Ali Farooq,~\IEEEmembership{Member,~IEEE, }%
Wang Yao, %
Peter Corcoran,~\IEEEmembership{Fellow,~IEEE}
\thanks{Muhammad Ali Farooq, Wang Yao, and Peter Corcoran are with the College of Science and Engineering, University of Galway, Galway, H91 TK33 Ireland.}
}

\markboth{Journal of \LaTeX\ Class Files,~Vol.~14, No.~8, August~2021}%
{Shell \MakeLowercase{\textit{et al.}}: A Sample Article Using IEEEtran.cls for IEEE Journals}

\IEEEpubid{0000--0000/00\$00.00~\copyright~2021 IEEE}

\maketitle

\begin{abstract}
In this research work we have proposed high-level ChildDiffusion framework capable of generating photorealistic child facial samples and further embedding several intelligent augmentations on child facial data using short text prompts, detailed textual guidance from LLMs, and further image to image transformation using text guidance control conditioning thus providing an opportunity to curate fully synthetic large scale child datasets 
The framework is validated by rendering high-quality child faces representing ethnicity data, micro expressions, face pose variations, eye blinking effects, facial accessories, different hair colours and styles, aging, multiple and different child gender subjects in a single frame. Addressing privacy concerns regarding child data acquisition requires a comprehensive approach that involves legal, ethical, and technological considerations. Keeping this in view this framework can be adapted to synthesise child facial data 
which can be effectively used for numerous downstream machine learning tasks. The proposed method circumvents common issues encountered in generative AI tools, such as temporal inconsistency and limited control over the rendered outputs. As an exemplary use case we have open-sourced child ethnicity data consisting of 2.5k child facial samples of five different classes which includes African, Asian, White, Indian, and Hispanic races by deploying the model in production inference phase. The rendered data undergoes rigorous qualitative as well as quantitative tests to cross validate its efficacy and further fine-tuning Yolo architecture for detecting and classifying child ethnicity as an exemplary downstream machine learning task.
\end{abstract}


\section{Introduction}
\begin{figure}[!thpb]
    \includegraphics[width=0.95\linewidth]{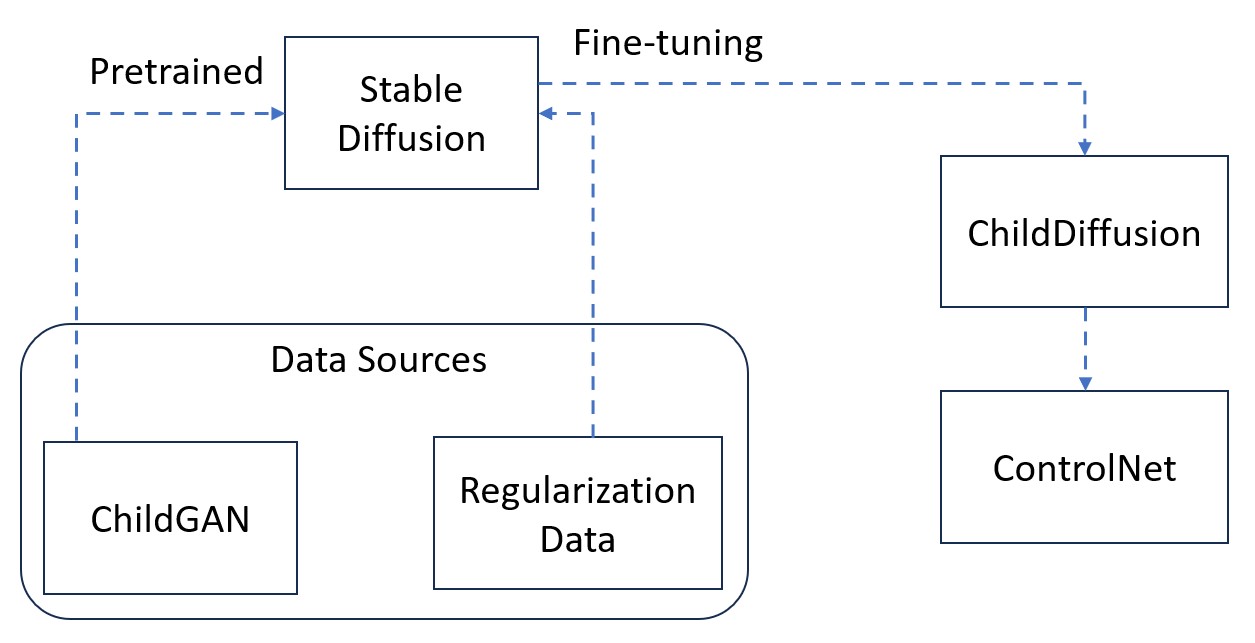}
    \caption{ChildDiffusion building methodology by employing pretrained Imagen architecture.}
    \label{fig_1}
\end{figure}

\begin{figure}[!thpb]
    \includegraphics[width=0.95\linewidth]{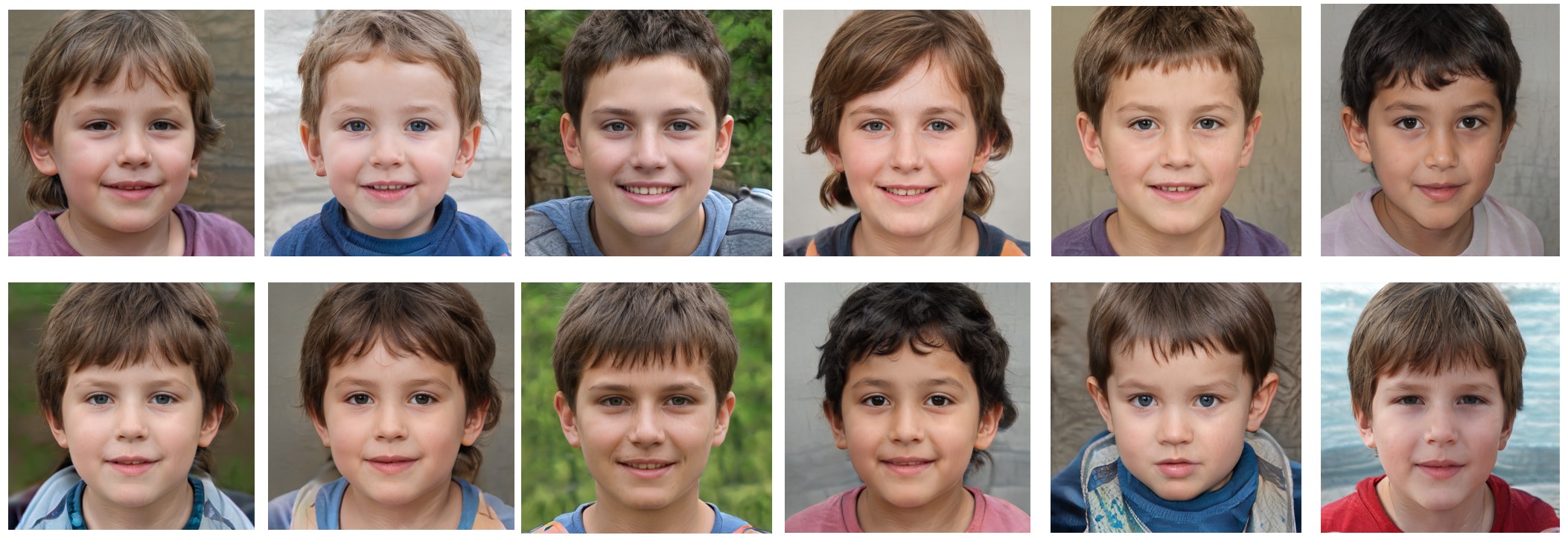}
    \includegraphics[width=0.95\linewidth]{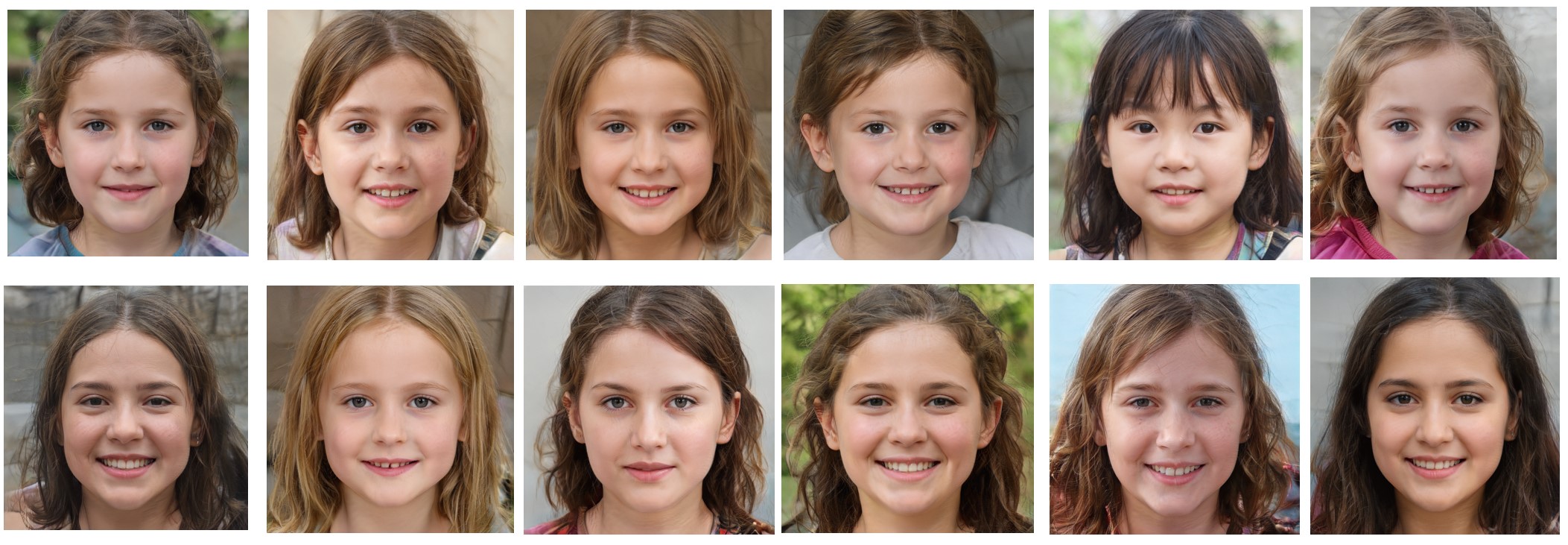}    
    \caption{ChildDiffusion tuning data acquired from ChildGAN generated outputs. The first two rows show the boy face samples whereas last two rows show the girl face samples.}
    \label{fig_2}
\end{figure}

\IEEEPARstart{W}{hen} coming towards acquiring child data It is essential to establish robust safeguards, regulatory frameworks, and ethical guidelines to protect children's privacy rights while promoting responsible data usage and innovation in healthcare, education, and technological domain. However practically sometimes it becomes difficult to acquire large scale child data with specific user requirements as per European GDPR for desired application. Synthetic data generation techniques offer flexibility in creating datasets tailored to specific use cases and requirements. Researchers and practitioners have control over the characteristics and properties of the synthetic data, allowing them to address specific challenges or explore challenging scenarios. This work is continuation phase of our previous research work where we have proposed ChildGAN architecture build upon StyleGAN2~\cite{karras2020analyzing} as base model to render photo realistic child facial data with six different facial transformations~\cite{farooq2023childgan}. However, due to the limited number of advanced augmentations and further certain limitations in GAN’s such that domain specificity where the model trained on specific datasets may struggle to generalize to new or unseen data distributions, sensitivity to hyperparameters such as minor changes in hyperparameters can significantly impact the training dynamics and the quality of generated samples. Despite of certain advances like employing SoA loss functions, GANs sometimes generate samples that are not of high fidelity, and variation derived us to propose a more robust text to image and image to image diffusion framework. This eventually allows to enhance the controllability of generated outputs and promote robustness and generalization across different dataset and domains. The recent succusses of latent diffusion models~\cite{rombach2022high} which represent a powerful advancement in the field of generative AI, particularly in the realm of image synthesis and data generation. These models leverage diffusion processes to transform a simple initial distribution (latent space) into complex, high-dimensional data distributions, such as images or audio waveforms. This process is controlled by diffusion steps, where each step corresponds to a specific amount of noise added to the latent space. Figure~\ref{fig_1} shows the streamlined ChildDiffusion building methodology where we have employed pretrained Imagen text to image (T2I) stable diffusion model~\cite{saharia2022photorealistic} developed by google brain team. The model sets a new benchmark in performance with a state-of-the-art FID score of 7.27 on the COCO dataset~\cite{lin2014microsoft}, accomplished without any prior training on COCO itself. The training data to tune ChildDiffusion model was acquired from ChildGAN rendered outputs as shown in Figure~\ref{fig_2}. The acquired data incorporates the boys and girl’s facial samples as depicted in Figure~\ref{fig_2} along with complete variety of smart facial transformations generated via ChildGAN architecture such that latent facial expressions, child aging progression, eye blinking effects, head pose variations, skin, and hair colour variation, and distinct directional lighting conditions.

\IEEEpubidadjcol

Training latent diffusion models typically involves maximizing the likelihood of the observed data given the latent variables. Despite their computational complexity, diffusion models offer several advantages, including the generation of high-fidelity data samples, controllable synthesis of data distributions, and improved stability compared to other generative models like Generative Adversarial Networks (GANs)~\cite{de2023review}. In addition to photo realistic image synthesis diffusion models have various applications, including, denoising~\cite{xia2023diffir}, upscaling image quality using super resolution~\cite{shang2024resdiff}, image editing~\cite{yang2024dynamic}, and data augmentation~\cite{dunlap2024diversify}. They have shown promising results in generating diverse, realistic data samples, making them a valuable tool in the machine learning domain. In this work we have used transfer learning methodology for fine-tuning pretrained Imagen architecture to render high-quality child facial data with countless possibilities to embed new transformations based on text prompts. Further we have also used ControlNet~\cite{zhang2023adding} that enables Stable Diffusion to take in a condition input that guides the image generation process, resulting in enhanced performance of Stable Diffusion models. It offers a variety of conditional inputs models which includes scribbles, edge maps, pose key points, tiles, depth maps, segmentation maps, normal maps, to guide the content of the generated image. In this way we can produce more comprehensive conditional transformations using text to image and image to image translation. The core contributions of this work are as follows.

\begin{itemize}
    \item Adapting and robust finetuning of stable diffusion model, with seed data generated from ChildGAN. Additionally, the proposal of weighted sum model merging techniques constitutes a significant advancement in the development of the Childdiffusion framework.
    \item Augmenting new child synthetic data with precise facial features based on textual guidance and open sourcing a novel child race synthetic data rendered via ChildDiffusion.
    \item The implementation of advanced data augmentation strategies, leveraging additional control-guided annotations and detailed text prompts extrcated via Large language models (LLM’S). These methodologies enrich the diversity and fidelity of synthetic child data, enhancing its utility in various applications.
    \item The comprehensive validation of synthetic image data quality by employing various qualitative and quantitative metrics to validate the efficacy of the proposed diffusion imaging framework.
\end{itemize}

\section{Background} \label{background}
Child face data plays an important role in various fields, such as tracing missing children through facial recognition and helping mental health issues by analyzing facial features and expressions. However, it is challenging to collect large-scale real-world diverse child data due to the General Data Policy Regulations (GDPR). Existing real-world children datasets such as YFA~\cite{bahmani2022face}, ITWCC~\cite{ricanek2015review}, NITL~\cite{best2016automatic}, CLF~\cite{deb2018longitudinal}, ICD~\cite{chandaliya2022childgan} and LCFW~\cite{jin2022double} are collected or web-scraped from children's faces. Most of them are private, and while some are available on request, they have strict rules for their use. The unique character of children's faces makes it difficult to collect and disclose this data. This raises the question of whether we can generate synthetic children's faces in the real world. 
Recently some research has explored generating synthetic child datasets by using GAN-based architectures. Falkenberg et al.~\cite{falkenberg2023child} proposed the HDA-SynChildFaces dataset which uses StyleGAN3~\cite{Karras2021} to generate adult data and progressed to children data by using InterFaceGAN~\cite{shen2020interfacegan} methods. Muhammad et al.~\cite{farooq2023childgan}  proposed the ChildGAN dataset which adopts transfer learning on the StyleGAN2~\cite{Karras2019stylegan2} network. However, these methods have limited flexibility and are not targeted at synthesizing ethnically diverse images of children. The emergence of text-driven diffusion-based modeling provides new solutions for more flexible image generation. For example, \cite{melzi2023gandiffface, banerjee2023identity} have used diffusion models to synthesize face images flexibly and efficiently. 

Text-driven image manipulation also known as text-to-image synthesis has recently made significant progress since the emergence of image diffusion models. Dhariwal et al.~\cite{NEURIPS2021_49ad23d1} have devised a series of iterative steps to progressively transform initial noise into the intended image. This approach preserves both diversity and richness in the generated content, showcasing the potential of diffusion models to surpass the image sample quality achieved by current state-of-the-art generative models. Rombach et al.~\cite{rombach2022high} propose Latent Diffusion Models (LDM) which reduce the computation cost in diffusion steps. 
The previous method works well with a long time of training, but in practical applications, we need customized models that have been tailored or adapted to meet specific requirements or address particular challenges in image generation tasks. Few-shot learning was employed, which uses limited samples to fine-tune a large-scale trained diffusion model. Hu et al.~\cite{hu2021lora} propose Low-Rank Adaptation (LoRA), which focuses on training only a subset of parameters using rank decomposition matrices. This approach significantly reduces computational and storage requirements while maintaining high performance. Ruiz et al.~\cite{ruiz2023dreambooth} propose DreamBooth, which fine-tunes a pre-trained text-to-image model with just a few images of a subject, enabling the model to learn to bind a unique identifier with that subject. Inspired by the strong transfer learning and few-shot learning abilities, we adopt LoRA and Dreambooth to train the ChildDiffusion network.

To provide personalized control over the synthetic image, several image diffusion process provides control on diffusion models. Zhang et al.~\cite{zhang2023adding} propose ControlNet, which could learn conditional controls for large diffusion models. ControlNet is a neural network architecture designed to work alongside large pre-trained text-to-image diffusion models. ControlNet acts as a sort of bridge between the conditioning inputs and the pretrained text-to-image diffusion model. It can be used to control stable diffusion and condition on Canny edges, Hough lines, user scribbles, human keypoints, segmentation maps, shape normals, depth, and cartoon line drawings. It does this by preserving the quality and capabilities of the large model while allowing for the learning of diverse conditional controls. The sampler is an important component in image generation like stabilizing diffusion.\footnote{\url{https://gilgam3sh.medium.com/tutorial-what-is-a-sampler-in-stable-diffusion-d5c16875b898}} Efficient exploration of parameters and reduction of sample variance can be achieved by employing various stabilizing diffusion samplers. This allows for more accurate estimation of the target distribution. Each sampler comes with its own set of advantages and disadvantages, making it crucial to select the appropriate one based on factors such as image complexity, resolution, and the type of cue being utilized.

In this work, we propose a new child data generation network, that mainly focuses on generating photo-realistic child facial data by integrating stable diffusion and few-shot learning. We have proposed a ChildDiffusion network that could generate child face data by inputting a simple text prompt. In addition, we have published a synthetic child dataset, which has diverse ethnicities with various accessories.

\section{Proposed Methodology}
\begin{figure*}[!thpb]
    \includegraphics[width=0.99\linewidth]{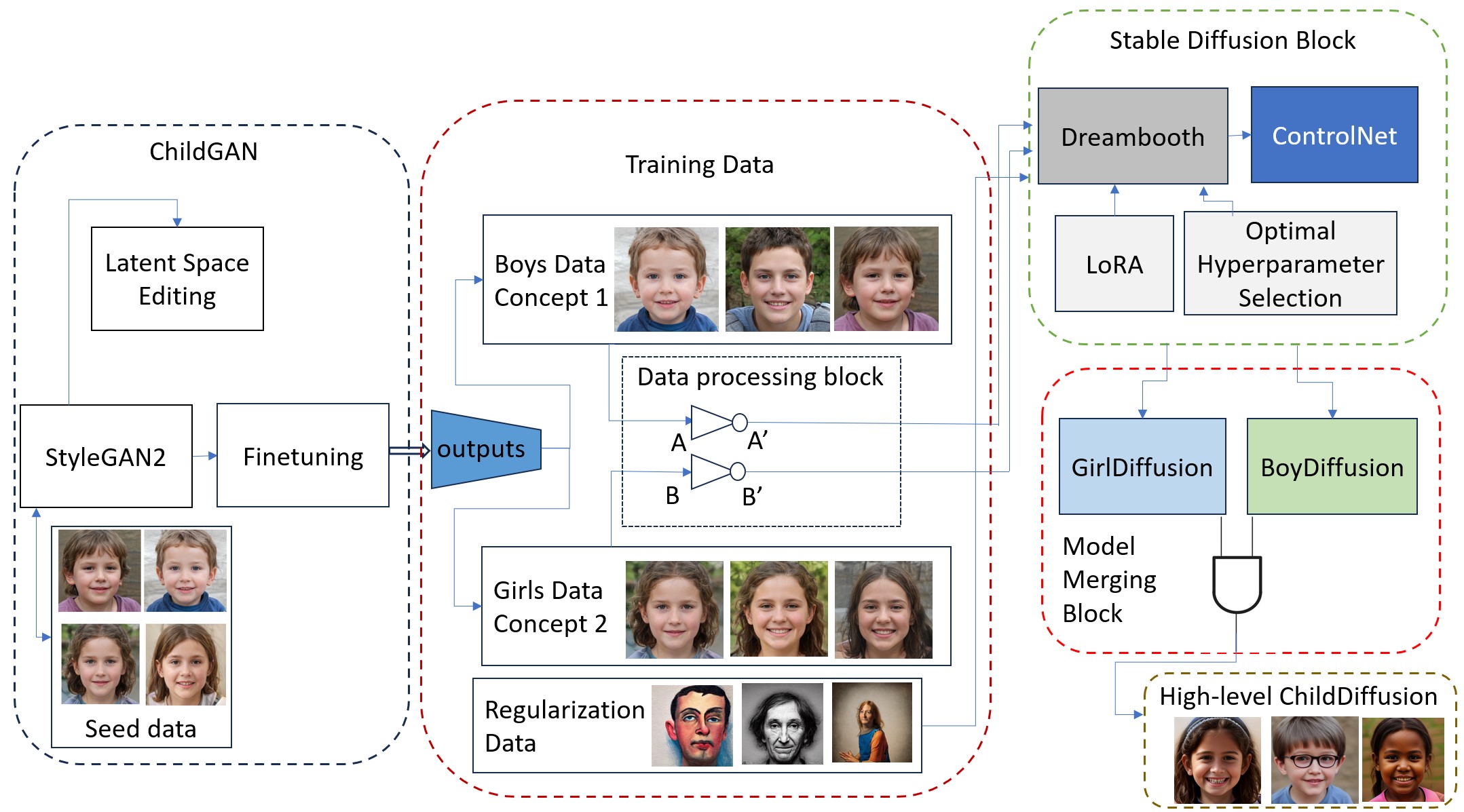}  
    \caption{Comprehensive block diagram representation for building high-level ChildDiffusion framework.}
    \label{fig_3}
\end{figure*}

This section will elaborate the adapted methodology employed in development the Childdiffusion framework. As mentioned in Section~\ref{background} we have incorporated transfer learning to tune pretrained stable diffusion models. Figure~\ref{fig_3} depicts the comprehensive block diagram representation for building Childdiffusion architecture and its 
subsequent utilization for the inference of high-quality child face data. 

\subsection{Data Sourcing} The first phase includes collecting training data along with regularization data for feeding it to stable diffusion data loaders. This was achieved through the ChildGAN network. The ChildGAN architecture ~\cite{farooq2023childgan}, derived from StyleGAN2, underwent refined convergence through domain adaptation and fine-tuning methodologies. This enhanced architecture was subsequently utilized to produce scalable children's faces, incorporating diverse smart transformations and attributes. Thus, taking the advantage of ChildGAN we gathered diversified seed data comprising of boys and girls faces along with all the possible smart transformations. The overall training data was divided in two different classes/ concepts which includes boys’ face data and girls face data along with respective tokens (descriptive words) that describe the data class. The acquired data underwent preprocessing to enhance its quality, including normalizing and resizing it to $512 \times 512$ pixels and removing any noise artifacts to ensure providing refined outputs to the data loaders. The reason for image resizing is due to the Stable Diffusion pretrained models are typically trained on $512 \times 512$ images and further trying to generate larger images can result in image distortion.

\begin{figure}[!h]
    \includegraphics[width=0.9\linewidth]{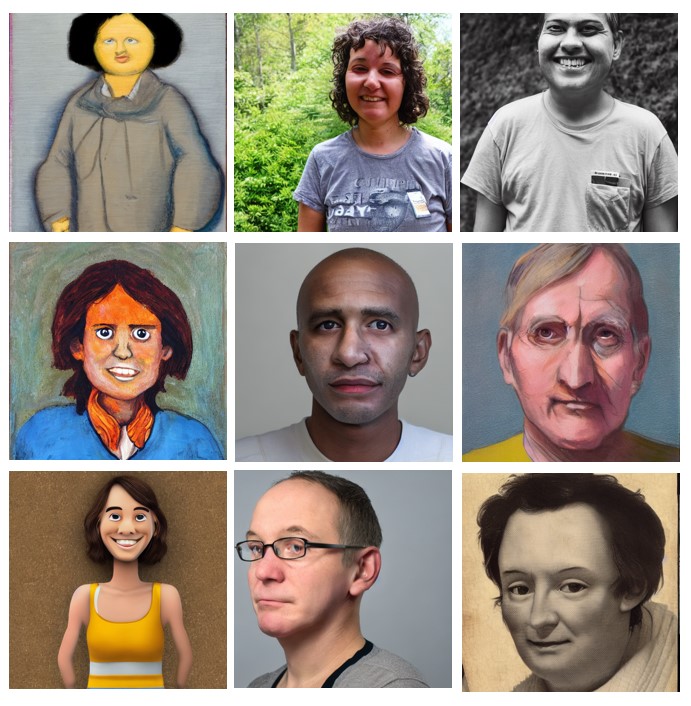}  
    \caption{Samples images of regularization data.}
    \label{fig_4}
\end{figure}

In addition to this we have incorporated a subset of person class regularization data in the training pipeline that are used as part of a regularization process to improve the stability and performance of pretrained stable diffusion models. Figure~\ref{fig_4} shows the sample images of regularization subset.

\subsection{Stable Diffusion Tuning Components} As illustrated in Figure~\ref{fig_3} we have used Dreambooth tool~\cite{ruiz2023dreambooth} for personalized text guided child face synthesis. It is a fine-tuning method that update the pretrained stable diffusion model weights by utilizing few-shot image learning depicting a specific subject or style. This technique operates by linking a unique keyword in the prompt with the provided example images. Through harnessing the inherent semantic understanding embedded within the pretrained Imagen model and incorporating a novel autogenous class-specific prior preservation loss, the adapted approach facilitates the synthesis of high-quality child facial data across different genders, poses, viewpoints, expressions, and lighting conditions which is typically not present in the initial training data. In addition to this we have shortlisted optimal set of network hypermeters after rigorous initial experimentation during the training phase. Two identical models BoyDiffusion and GirlDiffusion were trained using 8 bit, AdamW optimizer with base learning rate of $1.0e^{-04}$.  The models were trained using FP16 mixed precision~\cite{micikevicius2017mixed}. Mixed precision training provides a substantial computational speedup by performing operations in half-precision format, while conserving minimal information in single-precision storage to preserve crucial data integrity within key segments of the network. Further we have employed Low Rank Apatation (LoRA) model~\cite{hu2021lora} within UNet and text encoder which is a mathematical technique to reduce the number of model parameters rather than using the complete weights thus to minimize the computational costs. LoRA was created as a faster and more efficient way of fine-tuning large language models~\cite{liu2024alora}. Later the same concept was adapted for fine-tuning large diffusion models for various imaging applications~\cite{farooq2024derm, abaid2024synthesizing}. In our case the LoRA Unet learning rate was set to $1.0e^{-04}$ and LoRA text encoder learning rate was set to $5.0e^{-06}$.

\subsection{Model Merging} As depicted in Figure~\ref{fig_3}, block 4, two distinct personalized stable diffusion models were trained: boydiffusion and girldiffusion. The rationale behind training these models separately is to capture the optimal conceptual representations unique to each gender. To create a unified diffusion framework capable of executing all intelligent transformations, we employed a model merging technique. This approach involves combining both tuned models into a single block using weighted sum as shown in equation 1, which is then stored as a unified model. This amalgamation allows for seamless execution of diverse transformations for all the child genders within a single framework.

\textit{Model merging using weighted sum:} 
\begin{equation}
  (1-\alpha) \cdot A + \alpha \cdot B  
\end{equation}

Where A and B defines the model and $\alpha$ is used to assign weights to specific model. In our case we have selected $\alpha$ value to 0.6 after doing initial experimentation thus providing $40\%$ weightage to boydiffusion model and $60\%$ weightage to girldiffusion model.

\subsection{Smart Transformation via Additional Control} 
\begin{figure*}[!thpb]
    \includegraphics[width=0.99\linewidth]{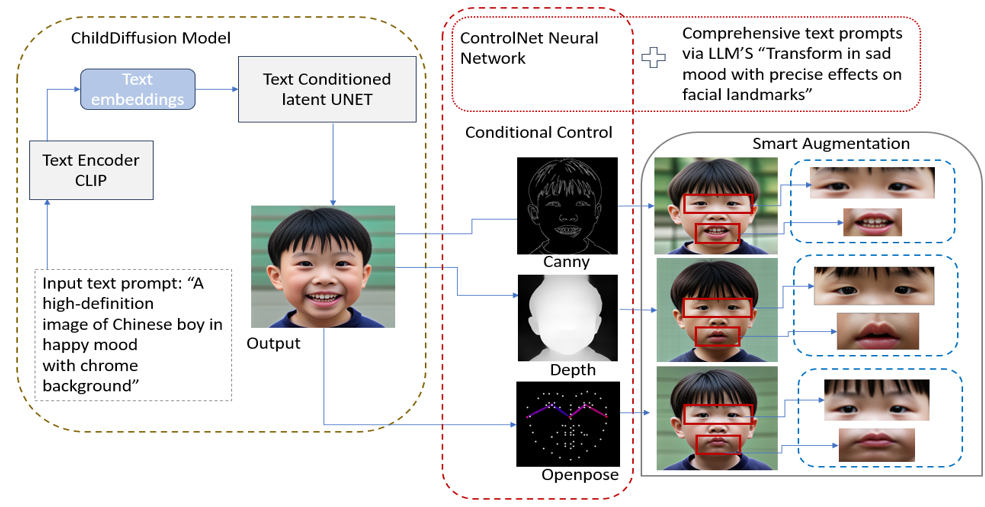}  
    \caption{Applying ControlNet annotations models (edges, depth, and open pose) for extracting spatial information to produce text guided facial expression on race specific child facial data.}
    \label{fig_5}
\end{figure*}

ControlNet is used as it plays pivotal to incorporate additional conditions for extensive transformations in rendered outputs generated via unified ChildDiffusion framework.  This framework advances and support for diverse spatial contexts, including edge detector, depth, and segmentation maps, open pose, scribbles, and key points, which can function as supplementary conditions for tuned text to image stable diffusion. Figure~\ref{fig_5} shows the functional overview of ControlNet conditions merged with ChildDiffuison framework for embedding complex transformations on rendered child facial data.

\section{Experimental Results}\label{exp}
The experimental environment was setup on workstation machine equipped with A6000 graphics card having 48 GPU of graphical memory interface.  Incorporating and enabling LoRA assists in reducing video memory footprints, leading to more efficient resource utilization. The training process for both models was completed in approximately 72 hours. The BoyDiffusion model required approximately 4.7 GB of VRAM, while the GirlDiffusion Model required approximately 4.6 GB of VRAM during fine-tuning phase. 

\subsection{Child Synthesis Inference Results using diverse Image Sampling Methods} 
\begin{figure}[!thpb]
    \includegraphics[width=0.98\linewidth]{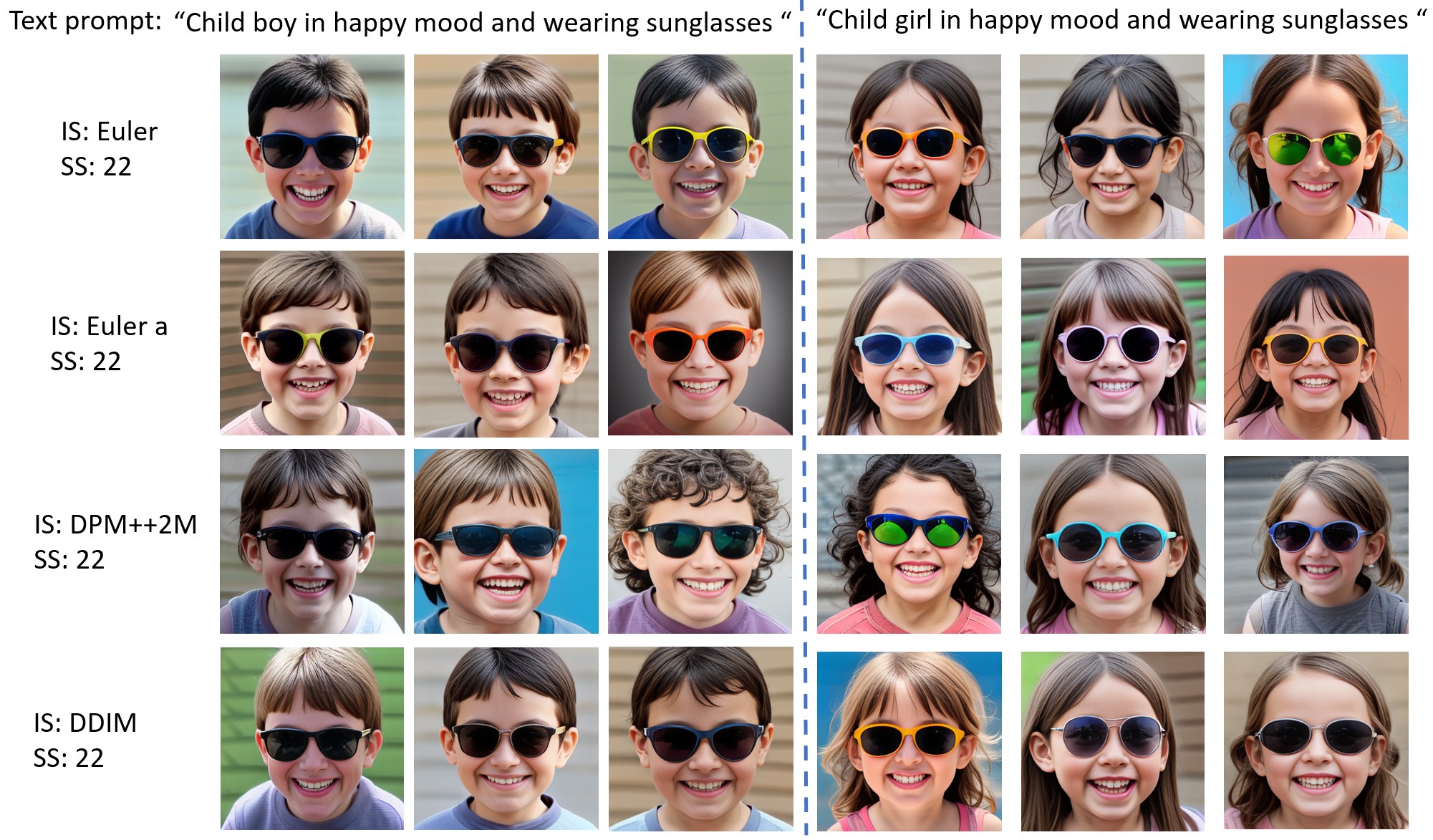}  
    \caption{Unique child faces with same expression by employing various SoA image sampling methods.}
    \label{fig_6}
\end{figure}

The trained models were deployed within the inference pipeline by employing diverse image sampling methods along with varying sampling steps to explore the quality of generated outputs. In stable diffusion image sampler operates by iterating through a given number of steps. At each step, it samples the latent space to calculate the local gradient, determining the direction for the subsequent step. In this research work we have used euler, euler A, DPM++ 2M, and DDIM sampling methods to explore the diversity in rendered data. Figure~\ref{fig_6} shows the rendered child facial expressions with facial accessories outputs using different sampling methods with same text guided prompt.

Figure~\ref{fig_6} illustrates that while each sampling method has demonstrated robust outcomes in generating photorealistic samples of both boy and girl faces, the DPM++ 2M approach exhibits a marginally superior performance. This is evidenced by its consistent ability to generate diversified samples, characterized by distinct variations in hairstyles, an array of sunglasses styles and shades, varied facial angles, and diverse facial structures.

\subsection{Text Guided Transformations} 
\begin{figure*}[!thpb]
\centering
    \includegraphics[width=\linewidth]{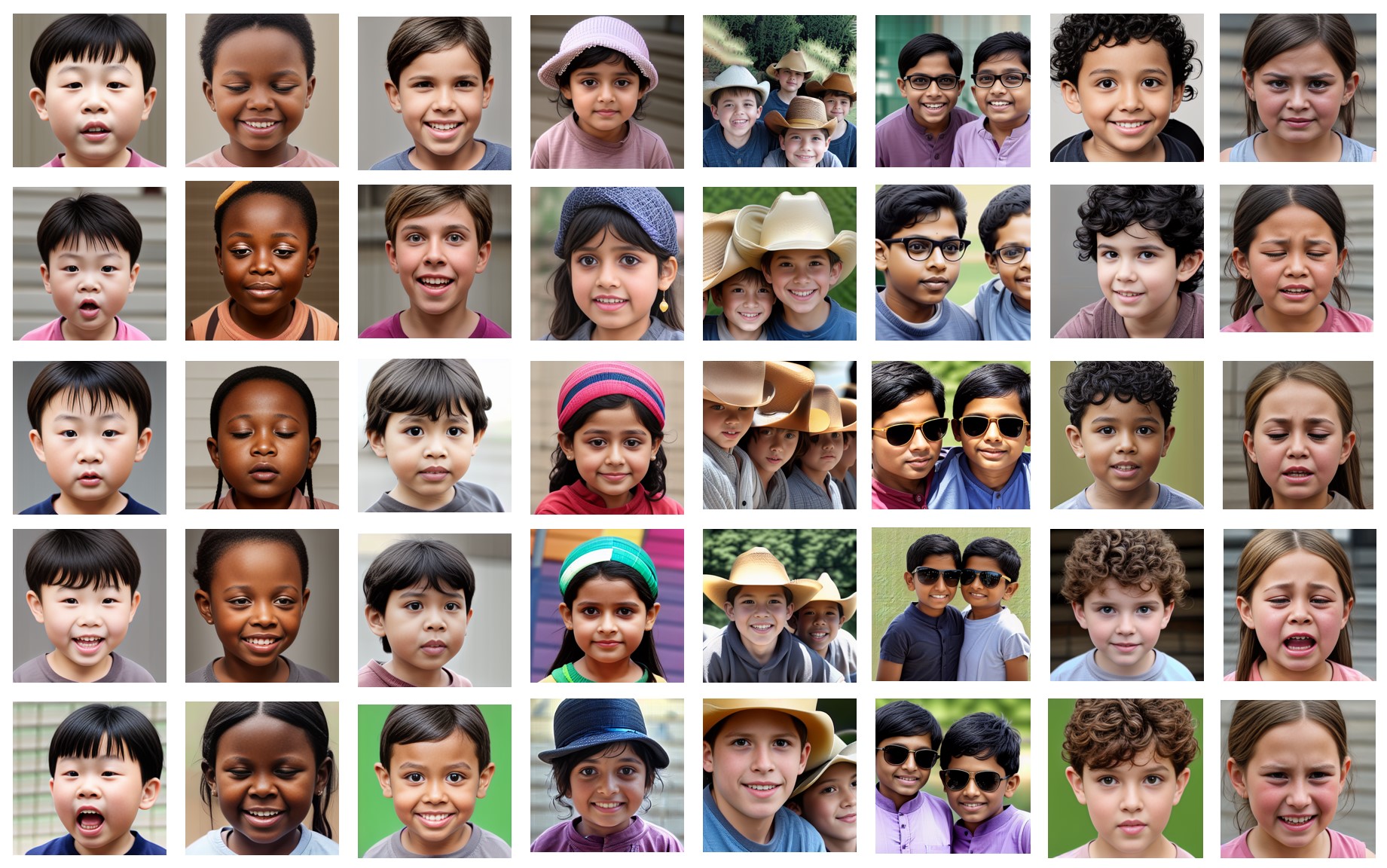}
    a \qquad \qquad \qquad b \quad \qquad \qquad c \qquad \qquad \qquad d \quad \qquad \qquad e \qquad \qquad \qquad f \qquad \qquad \qquad g  \qquad \qquad \qquad h \\
    \caption{Unique child faces with various smart transformations by providing text guided prompts to Childdiffusion model, a) text prompt: Chinese boy with unique facial expressions, b) tp: African girl with eye blinking and eye closing effects, c) tp: different lighting conditions and head pose variations embedded on child face, d) tp: Pakistani girl wearing different types of hats, e) tp: group of boys wearing cow boy hats, f) pair of Indian boys wearing glasses and sunglasses, g) boy with black and brown curly hairs, h) tp: child girl stressed and crying.}
    \label{fig_7}
\end{figure*}

Stable Diffusion leverages CLIP~\cite{radford2021learning}, an innovative model renowned for its capacity to assimilate diverse data types and execute a broad spectrum of tasks. The CLIP embedding process encompasses multiple layers, progressively enhancing the depth and intricacy of the representation. Consequently, a more elaborate prompt yields a more robust data representation. In this section, we introduce intelligent transformations facilitated by text-guided prompts applied to child facial data, thereby exploring the potential of the proposed ChildDiffusion Model. Figure~\ref{fig_7} demonstrates various smart transformations rendered via Childdiffusion model by employing all the image sampling techniques and shortlisting the best results based on visual inspections. Further specific text prompts are used as inputs to embed that type of effects on child facial data.

Figure~\ref{fig_7} demonstrates that the ChildDiffusion framework accelerates various sophisticated data transformations contingent upon user-specified textual prompts. Further it took appx 10 seconds to generate images with batch size of 8 and 1 batch count. This capability can be expanded by providing distinct text prompts tailored to meet user-specific needs. It is important to mention that we have not used any type of clip skip operation at this stage.  CLIP Skip facilitates the omission of certain layers, thereby mitigating computational requirements and expediting the generation process. This feature proves especially beneficial for handling intricate prompts necessitating substantial processing power and time. However, clip skip feature can be used when providing more detailed text prompts. 

\subsection{Additional styles and control in ChildDiffsuion via ControlNet:}
\begin{figure*}[!thpb]
    \includegraphics[width= 0.99\linewidth]{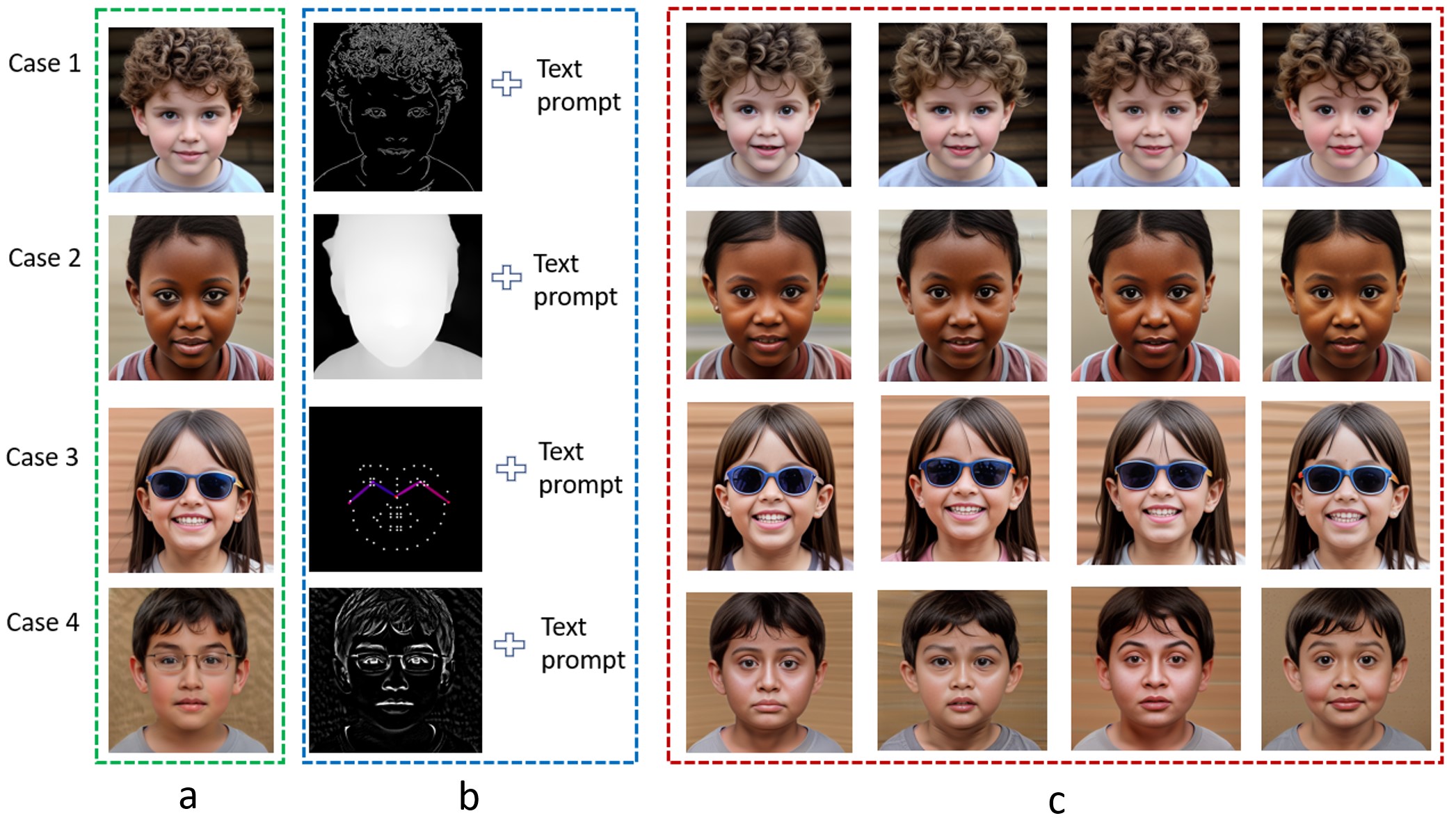}
    \caption{Controlling ChildDiffusion model with various Conditions and detailed text prompts to achieve smart augmentation on 4 different text cases, a) input images/ generated via Childdiffusion model, b) provided specific textual guidance prompt as provided in Table~\ref{tab1} along with ControlNet conditioning which includes edge map for case 1, depth map for case 2, openpose for case 3 and lineart for case 4, c) resulting transformations outputs rendered via ChildDiffusion framework.}
    \label{fig_8}
\end{figure*}

In addition to capabilities of ChildDiffusion model explored in earlier sections we have further discovered the robustness of proposed network by providing more detailed textual prompts generated via large language models (LLMs) as the input conditions (x0) along with additional controlled conditions (y) through ControlNet architecture. This in turn provides spatial consistency for various types of image outputs rendered via stable diffusion models. This phenomenon is exemplified in Figure~\ref{fig_8}, wherein the employment of more refined textual prompts alongside ControlNet pretrained detectors facilitated the inference of various facial effects across four distinct test cases. Table~\ref{tab1} shows the specific text prompt for each of the test case.

\begin{table*}[!thpb]
\centering
\caption{Textual Prompts for embedding special effects on child facial data.}
\label{tab1}
\resizebox{\textwidth}{!}{%
\begin{tabular}{|c|l|}
\hline
Case No & \multicolumn{1}{c|}{Textual guidance prompt extracted from LLM's} \\ \hline
1 &
  \begin{tabular}[c]{@{}l@{}}“Generate a series of images depicting the child's face transitioning through a range of \\ emotions, such as happiness, sadness, anger, surprise, and fear. Ensure the transformations \\ are nuanced and realistic, capturing subtle changes in muscle movement, eye gaze, and \\ mouth shape.”\end{tabular} \\ \hline
2 &
  \begin{tabular}[c]{@{}l@{}}“Develop a transformation to investigate the impact of environmental factors on facial \\ morphology. Generate variations in the child's facial features, such as skin tone, facial \\ structure, and facial expressions, to simulate exposure to different climates, diets, and \\ lifestyles.\end{tabular} \\ \hline
3 &
  \begin{tabular}[c]{@{}l@{}}“Create an advanced transformation to explore the impact of hairstyle modifications on \\ facial perception and identity. Generate facial images and apply sophisticated hairstyle \\ manipulation techniques to experiment with a variety of haircuts, colours, and styling \\ options.”\end{tabular} \\ \hline
4 &
  \begin{tabular}[c]{@{}l@{}}“Remove the glasses and further simulate the effects of facial expression dynamics on cheek\\  contour and prominence. Generate facial images displaying a range of facial expressions, \\ such as smiling, frowning, and pouting.”\end{tabular} \\ \hline
\end{tabular}%
}
\end{table*}

\section{Synthetic Child Race Data}
\begin{figure*}[!thpb]
\centering
    \includegraphics[width=\linewidth]{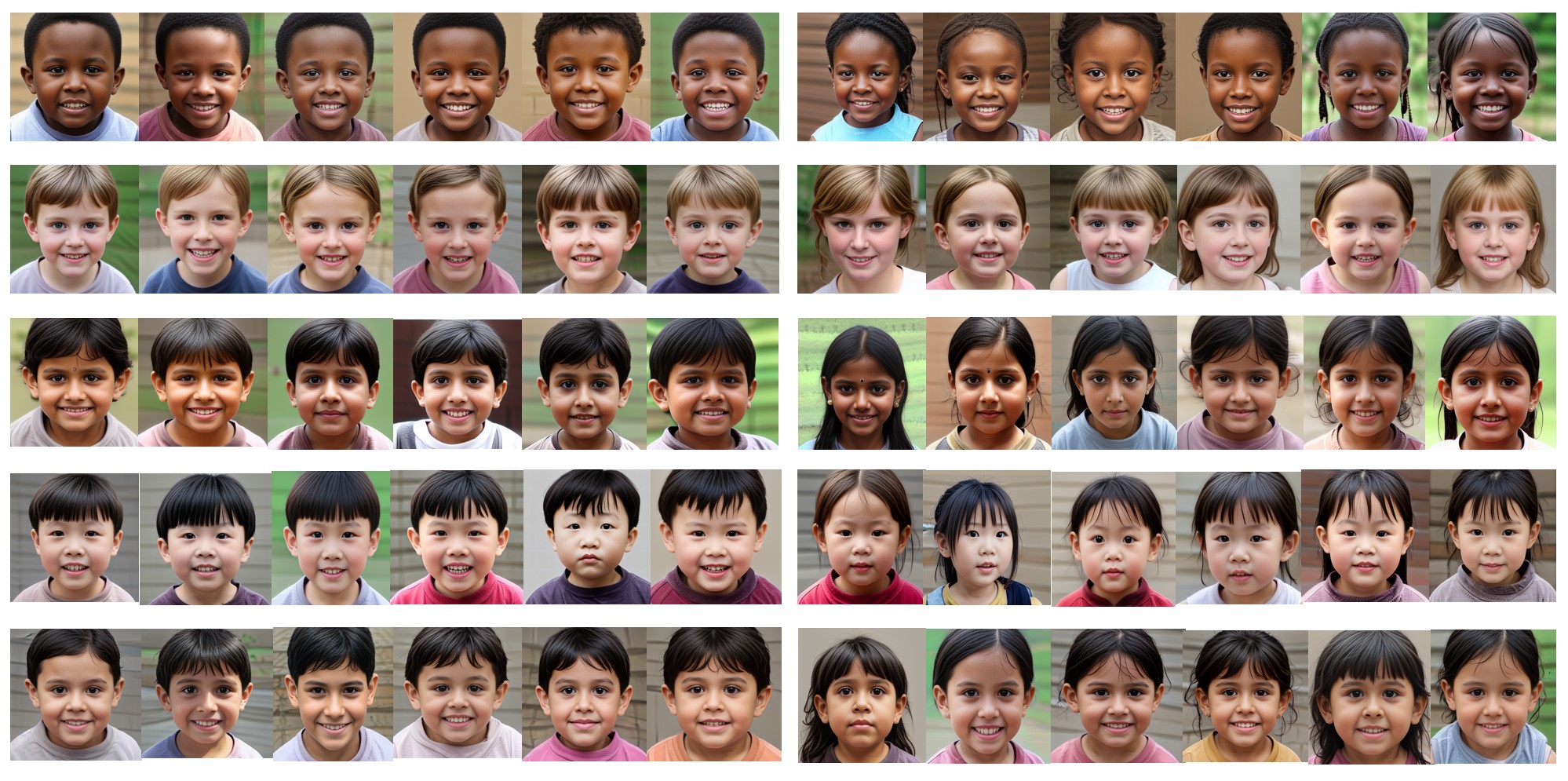}
    \caption{Child race data generated via ChildDiffusion model, the first column shows the African boys’ and girls’ samples, the second rows depict the American boys and girls samples, third row shows the Indian child data samples, fourth rows shows Asian (Chinese) boys and girls face samples whereas the last row shows Hispanic child facial samples.}
    \label{fig_9}
\end{figure*}

Further in this research work, we have successfully generated extensive datasets representing children of diverse racial backgrounds, exemplifying a notable application of this childdiffusion framework and an exemplary use case. By harnessing the capabilities of latent diffusion models, we have been able to produce high-quality synthetic data that accurately reflects the inherent diversity within child populations across various ethnicities. This exemplar employment underscores the efficacy of Stable Diffusion in synthesizing large-scale datasets tailored to specific demographic characteristics, thereby serving as a pivotal resource for research and development endeavours in fields such as child psychology, education, and public health. This is accomplished by deploying the childdiffusion model in batched inference pipeline. The overall datasets consist of $2,500$ samples of boys and girls from five different race classes The complete dataset attributes are summarized in Table~\ref{tab2} whereas Figure~\ref{fig_9} showcases the child race facial image data from five distinct classes, highlighting the diverse facial characteristics represented within each class. The images exhibit robust quality, with clear delineation of facial features, expressions and face pose variations, demonstrating the efficacy of the proposed ChildDiffusion framework in generating high-fidelity by employing four different sampling methods discussed in Section~\ref{exp} part 1. 

\begin{table}[!thpb]
\centering
\caption{Synthetic Child Race Dataset Attributes.}
\label{tab2}
\resizebox{\linewidth}{!}{
\begin{tabular}{|c|c|c|c|c|}
\hline
Race Class & Total No. of Samples & Boys Data & Girls Data & \begin{tabular}[c]{@{}c@{}}Image Resolution \\ and storing format\end{tabular} \\ \hline
African  & 500 & 250 & 250 & \multirow{5}{*}{512x512, PNG} \\ \cline{1-4}
White    & 500 & 250 & 250 &                               \\ \cline{1-4}
Asian    & 500 & 250 & 250 &                               \\ \cline{1-4}
Indian   & 500 & 250 & 250 &                               \\ \cline{1-4}
Hispanic & 500 & 250 & 250 &                               \\ \hline
\end{tabular}%
}
\end{table}

At this point we have open-sourced child race dataset and Childdiffusion model via huggingface (Link). 
Further the same diffusion framework facilitates the generation of additional datasets according to user specifications, thereby offering flexibility and adaptability to meet diverse research or application needs.

\section{Synthetic Data Validation}
 Validating synthetic data generated for children requires careful consideration of several factors to ensure its accuracy, diversity, authenticity, and applicability in various contexts. By addressing these considerations and conducting thorough validation experiments, we can ensure the integrity, reliability, and utility of synthetic data in research and practical applications. Keeping this in view we have employed various validation tests, data evaluation metrics and data visualization tools as mentioned below.

\subsection{CLIP Score}
The CLIP score~\cite{taited2023CLIPScore} serves as a well-established metric for quantifying the similarity of an image to a given text. Utilizing such indicators is essential for assessing the quality of an image to ensure its alignment with a corresponding text caption. The Clip score is computed by calculating the dot product between the feature vectors of the image and the text, normalized to a range of [-1, 1]. Mathematically, the CLIP score is expressed as.

\textit{CLIP Score:} 
\begin{equation}
  \mathbf{normalize} (\textit{image\_feature\_vector} \cdot \textit{text\_feature\_vector}) 
\end{equation}

Where $\mathbf{normalize}$ represents the normalization function, \textit{image\_feature\_vector} denotes the feature vector extracted from the image using a pre-trained image encoder and \textit{text\_feature\_vector} represents the feature vector obtained from the text description using a pre-trained text encoder.

\begin{table}[!thpb]
\centering
\caption{The results of Clip Scores.}
\label{tab3}
\resizebox{\linewidth}{!}{
\begin{tabular}{c|c|c|c|c|c}
\hline \hline
     & DDIM  & DPM++2M & Euler & Euler a & Combine (Average)\\ \hline
Boy  & 34.76 & 34.22   & 34.23 & 34.50   & 34.42   \\ \hline
Girl & 35.11 & 34.68   & 35.05 & 35.09   & 34.98   \\ \hline \hline
\end{tabular}
}
\end{table}
Table~\ref{tab3} shows the CLIP score
results by collecting 100 distinct generated images of boys and girl’s class along with there corresponding text prompts by using all the image sampling methods as demonstrated in Figure~\ref{fig_6}. The results shows that the image results generated by employing DDIM image sampler perform best in comparison to other three sampling methods. Further it can be observed that the generated girl images have better CLIP Score than the boy images.

\subsection{Evaluation Metrics}
In addition to the CLIP score, we have computed three different accuracy metrics broadly used for evaluating the quality and diversity of rendered images via
diffusion generative models. This includes Frechet Inception Distance (FID)~\cite{heusel2017gans}, Inception Score (IS)~\cite{salimans2016improved}, and Kernal Inception Distance (KID)~\cite{bińkowski2018demystifying}. The quantitative results of these metrics are summarized in Table~\ref{tab4}. The majority of results demonstrates that rendered data comprises robust image quality and diversity. However it can be observed that in case of boys samples FID score is much higher as compared to girls. This might be due to image compression techniques for storing the data that hardly degrade the perceptual quality but induce high variations in the FID score \cite{otani2023toward}. Further it is noteworthy to mention that these scores were generated by race child data which has limited diversity on other attributes when compared to seed data samples generated via ChildGAN models which includes all the possible transformations including expressions, aging, lighting, eye blinking, and head pose thus diversity between two datasets are different.
\begin{table}[!thpb]
\centering
\caption{Evaluation Results.}
\label{tab4}
\resizebox{0.7\linewidth}{!}{
\begin{tabular}{c|c|c|c}
\hline \hline
    & Boy   & Girl  & Combine \\ \hline
FID $\downarrow$ & 56.23 & 36.99 & 39.07        \\ \hline
IS  $\uparrow$ & 2.15  & 1.75  & 2.33    \\ \hline
KID $\downarrow$ & 0.061 & 0.215 & 0.096   \\ \hline \hline
\end{tabular}
}
\end{table}

\begin{figure}[!thpb]
\centering
    \includegraphics[width=0.45\linewidth]{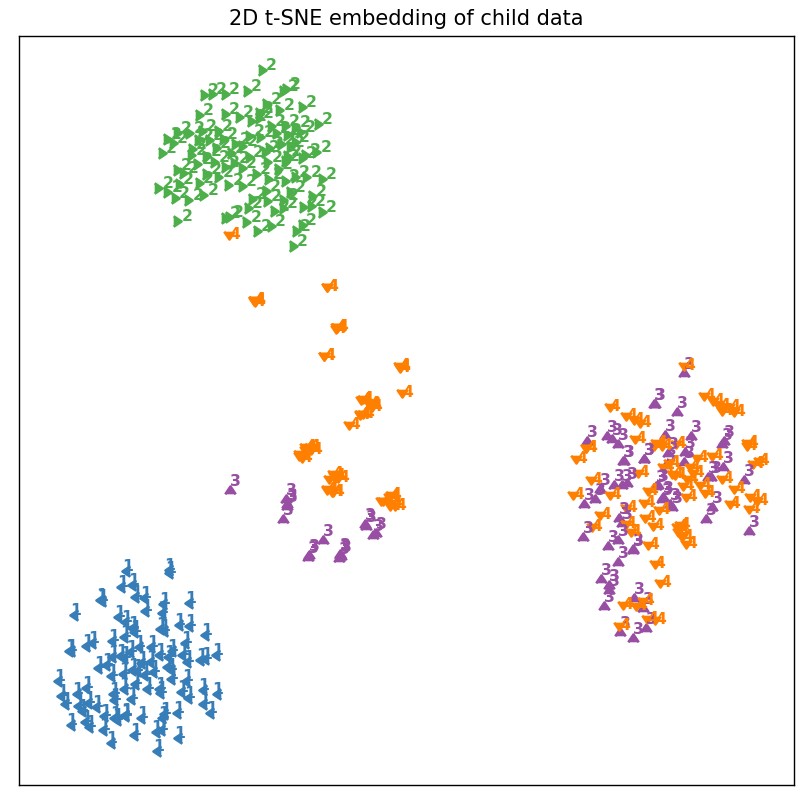} \quad 
    \includegraphics[width=0.48\linewidth]{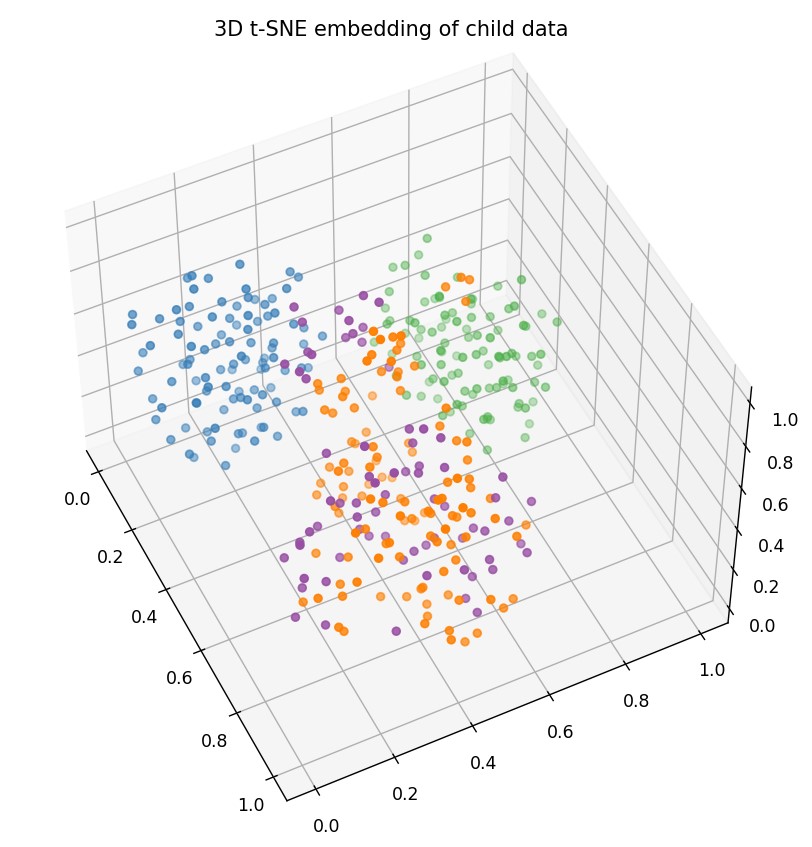}  
    \\  (a) \qquad \qquad \qquad \qquad \qquad \quad (b)
    \caption{t-SNE Visualization. Note blue denotes ChildDiffusion Boy, green denotes ChildDiffusion Girl, purple denotes Databrary Boy, and orange denotes Databrary Girl.}
    \label{fig_10}
\end{figure}

\subsection{tSNE Visualization}

t-SNE (t-distributed Stochastic Neighbor Embedding) \cite{van2008visualizing} is an unsupervised technique employed for non-linear dimensionality reduction, facilitating data exploration and visualization of high-dimensional datasets. This method enables the separation of data points that cannot be distinguished by linear boundaries, thereby enhancing the discernibility of complex relationships within the data. In this research work we have used t-SNE to visualize the real vs synthetic child data samples in both 2D and 3D maps as depicted in Figure~\ref{fig_10}. This is done by selecting a subset of 100 samples of each class i.e. boy and girls with different facial expressions generated via ChildDiffusion model with and further selecting real world child data samples from the Child Affective Facial Expression (CAFE) Databurry~\cite{10.3389/fpsyg.2014.01532} dataset. The CAFE dataset comprises a series of images capturing the facial expressions of children aged 2 to 8 years (Mean = 5.3 years; Range = 2.7 – 8.7 years). These images depict six emotional facial expressions: sadness, happiness, surprise, anger, disgust, fear, alongside a neutral expression. The complete dataset includes 90 female models and 64 male models, representing diverse racial backgrounds, including 27 African American, 16 Asian, 77 Caucasian/European American, 23 Latino, and 11 South Asian individuals. 

\section{Conclusion and Future work}
Overall, the findings presented in this study highlight the significant potential of the ChildDiffusion framework in addressing challenges associated with data scarcity and diversity in research and application contexts. In conclusion, the research undertaken has demonstrated the effectiveness and versatility of the proposed ChildDiffusion framework in generating synthetic child facial image data with robust quality. By leveraging advanced methodologies such as weighted model merging and precise facial feature augmentation guided by textual prompts, the framework showcases its ability to produce diverse datasets tailored to specific user requirements. Additionally, the integration of control-guided annotations and specific text control enhances the fidelity and utility of the synthetic data generated.

As the possible future directions, we can explore ways to incorporate domain-specific knowledge into the generation process to ensure that the synthetic data generated by the framework accurately reflects real-world scenarios and applications. Furthermore, Extend the evaluation of the ChildDiffusion framework to diverse computer vision applications, such as face recognition, emotion recognition, and demographic analysis. Lastly development of user-friendly graphic user interface and tools to streamline the usage of the ChildDiffusion framework thus to make it more accessible to a wider audience of researchers and practitioners.

\section*{Acknowledgments}
This work is supported by ADAPT - Centre for Digital Content Technology and Enterprise Ireland, and Irish Research Council Enterprise Partnership Ph.D. Scheme under Grant EPSPG/2020/40.

\bibliographystyle{IEEEtran}
\bibliography{ref.bib}

\begin{thebibliography}{10}
\providecommand{\url}[1]{#1}
\csname url@samestyle\endcsname
\providecommand{\newblock}{\relax}
\providecommand{\bibinfo}[2]{#2}
\providecommand{\BIBentrySTDinterwordspacing}{\spaceskip=0pt\relax}
\providecommand{\BIBentryALTinterwordstretchfactor}{4}
\providecommand{\BIBentryALTinterwordspacing}{\spaceskip=\fontdimen2\font plus
\BIBentryALTinterwordstretchfactor\fontdimen3\font minus \fontdimen4\font\relax}
\providecommand{\BIBforeignlanguage}[2]{{%
\expandafter\ifx\csname l@#1\endcsname\relax
\typeout{** WARNING: IEEEtran.bst: No hyphenation pattern has been}%
\typeout{** loaded for the language `#1'. Using the pattern for}%
\typeout{** the default language instead.}%
\else
\language=\csname l@#1\endcsname
\fi
#2}}
\providecommand{\BIBdecl}{\relax}
\BIBdecl

\bibitem{karras2020analyzing}
T.~Karras, S.~Laine, M.~Aittala, J.~Hellsten, J.~Lehtinen, and T.~Aila, ``Analyzing and improving the image quality of stylegan,'' in \emph{Proceedings of the IEEE/CVF conference on computer vision and pattern recognition}, 2020, pp. 8110--8119.

\bibitem{farooq2023childgan}
M.~A. Farooq, W.~Yao, G.~Costache, and P.~Corcoran, ``Childgan: Large scale synthetic child facial data using domain adaptation in stylegan,'' \emph{IEEE Access}, 2023.

\bibitem{rombach2022high}
R.~Rombach, A.~Blattmann, D.~Lorenz, P.~Esser, and B.~Ommer, ``High-resolution image synthesis with latent diffusion models,'' in \emph{Proceedings of the IEEE/CVF conference on computer vision and pattern recognition}, 2022, pp. 10\,684--10\,695.

\bibitem{saharia2022photorealistic}
C.~Saharia, W.~Chan, S.~Saxena, L.~Li, J.~Whang, E.~L. Denton, K.~Ghasemipour, R.~Gontijo~Lopes, B.~Karagol~Ayan, T.~Salimans \emph{et~al.}, ``Photorealistic text-to-image diffusion models with deep language understanding,'' \emph{Advances in neural information processing systems}, vol.~35, pp. 36\,479--36\,494, 2022.

\bibitem{lin2014microsoft}
T.-Y. Lin, M.~Maire, S.~Belongie, J.~Hays, P.~Perona, D.~Ramanan, P.~Doll{\'a}r, and C.~L. Zitnick, ``Microsoft coco: Common objects in context,'' in \emph{Computer Vision--ECCV 2014: 13th European Conference, Zurich, Switzerland, September 6-12, 2014, Proceedings, Part V 13}.\hskip 1em plus 0.5em minus 0.4em\relax Springer, 2014, pp. 740--755.

\bibitem{de2023review}
V.~L.~T. de~Souza, B.~A.~D. Marques, H.~C. Batagelo, and J.~P. Gois, ``A review on generative adversarial networks for image generation,'' \emph{Computers \& Graphics}, 2023.

\bibitem{xia2023diffir}
B.~Xia, Y.~Zhang, S.~Wang, Y.~Wang, X.~Wu, Y.~Tian, W.~Yang, and L.~Van~Gool, ``Diffir: Efficient diffusion model for image restoration,'' in \emph{Proceedings of the IEEE/CVF International Conference on Computer Vision}, 2023, pp. 13\,095--13\,105.

\bibitem{shang2024resdiff}
S.~Shang, Z.~Shan, G.~Liu, L.~Wang, X.~Wang, Z.~Zhang, and J.~Zhang, ``Resdiff: Combining cnn and diffusion model for image super-resolution,'' in \emph{Proceedings of the AAAI Conference on Artificial Intelligence}, vol.~38, no.~8, 2024, pp. 8975--8983.

\bibitem{yang2024dynamic}
F.~Yang, S.~Yang, M.~A. Butt, J.~van~de Weijer \emph{et~al.}, ``Dynamic prompt learning: Addressing cross-attention leakage for text-based image editing,'' \emph{Advances in Neural Information Processing Systems}, vol.~36, 2024.

\bibitem{dunlap2024diversify}
L.~Dunlap, A.~Umino, H.~Zhang, J.~Yang, J.~E. Gonzalez, and T.~Darrell, ``Diversify your vision datasets with automatic diffusion-based augmentation,'' \emph{Advances in Neural Information Processing Systems}, vol.~36, 2024.

\bibitem{zhang2023adding}
L.~Zhang, A.~Rao, and M.~Agrawala, ``Adding conditional control to text-to-image diffusion models,'' in \emph{Proceedings of the IEEE/CVF International Conference on Computer Vision}, 2023, pp. 3836--3847.

\bibitem{bahmani2022face}
K.~Bahmani and S.~Schuckers, ``Face recognition in children: a longitudinal study,'' in \emph{2022 International Workshop on Biometrics and Forensics (IWBF)}.\hskip 1em plus 0.5em minus 0.4em\relax IEEE, 2022, pp. 1--6.

\bibitem{ricanek2015review}
K.~Ricanek, S.~Bhardwaj, and M.~Sodomsky, ``A review of face recognition against longitudinal child faces,'' 2015.

\bibitem{best2016automatic}
L.~Best-Rowden, Y.~Hoole, and A.~Jain, ``Automatic face recognition of newborns, infants, and toddlers: A longitudinal evaluation,'' in \emph{2016 International Conference of the Biometrics Special Interest Group (BIOSIG)}.\hskip 1em plus 0.5em minus 0.4em\relax IEEE, 2016, pp. 1--8.

\bibitem{deb2018longitudinal}
D.~Deb, N.~Nain, and A.~K. Jain, ``Longitudinal study of child face recognition,'' in \emph{2018 International Conference on Biometrics (ICB)}.\hskip 1em plus 0.5em minus 0.4em\relax IEEE, 2018, pp. 225--232.

\bibitem{chandaliya2022childgan}
P.~K. Chandaliya and N.~Nain, ``Childgan: Face aging and rejuvenation to find missing children,'' \emph{Pattern Recognition}, vol. 129, p. 108761, 2022.

\bibitem{jin2022double}
X.~Jin, J.~Lei, S.~Ge, C.~Song, H.~Yu, and C.~Wu, ``Double-blinded finder: a two-side secure children face recognition system,'' \emph{Wireless Networks}, pp. 1--10, 2022.

\bibitem{falkenberg2023child}
M.~Falkenberg, A.~B. Ottsen, M.~Ibsen, and C.~Rathgeb, ``Child face recognition at scale: Synthetic data generation and performance benchmark,'' \emph{arXiv preprint arXiv:2304.11685}, 2023.

\bibitem{Karras2021}
T.~Karras, M.~Aittala, S.~Laine, E.~H\"ark\"onen, J.~Hellsten, J.~Lehtinen, and T.~Aila, ``Alias-free generative adversarial networks,'' in \emph{Proc. NeurIPS}, 2021.

\bibitem{shen2020interfacegan}
Y.~Shen, C.~Yang, X.~Tang, and B.~Zhou, ``Interfacegan: Interpreting the disentangled face representation learned by gans,'' \emph{TPAMI}, 2020.

\bibitem{Karras2019stylegan2}
T.~Karras, S.~Laine, M.~Aittala, J.~Hellsten, J.~Lehtinen, and T.~Aila, ``Analyzing and improving the image quality of {StyleGAN},'' in \emph{Proc. CVPR}, 2020.

\bibitem{melzi2023gandiffface}
P.~Melzi, C.~Rathgeb, R.~Tolosana, R.~Vera-Rodriguez, D.~Lawatsch, F.~Domin, and M.~Schaubert, ``Gandiffface: Controllable generation of synthetic datasets for face recognition with realistic variations,'' in \emph{Proceedings of the IEEE/CVF International Conference on Computer Vision}, 2023, pp. 3086--3095.

\bibitem{banerjee2023identity}
S.~Banerjee, G.~Mittal, A.~Joshi, C.~Hegde, and N.~Memon, ``Identity-preserving aging of face images via latent diffusion models,'' in \emph{2023 IEEE International Joint Conference on Biometrics (IJCB)}.\hskip 1em plus 0.5em minus 0.4em\relax IEEE, 2023, pp. 1--10.

\bibitem{NEURIPS2021_49ad23d1}
\BIBentryALTinterwordspacing
P.~Dhariwal and A.~Nichol, ``Diffusion models beat gans on image synthesis,'' in \emph{Advances in Neural Information Processing Systems}, M.~Ranzato, A.~Beygelzimer, Y.~Dauphin, P.~Liang, and J.~W. Vaughan, Eds., vol.~34.\hskip 1em plus 0.5em minus 0.4em\relax Curran Associates, Inc., 2021, pp. 8780--8794. [Online]. Available: \url{https://proceedings.neurips.cc/paper_files/paper/2021/file/49ad23d1ec9fa4bd8d77d02681df5cfa-Paper.pdf}
\BIBentrySTDinterwordspacing

\bibitem{hu2021lora}
E.~J. Hu, Y.~Shen, P.~Wallis, Z.~Allen-Zhu, Y.~Li, S.~Wang, L.~Wang, and W.~Chen, ``Lora: Low-rank adaptation of large language models,'' \emph{arXiv preprint arXiv:2106.09685}, 2021.

\bibitem{ruiz2023dreambooth}
N.~Ruiz, Y.~Li, V.~Jampani, Y.~Pritch, M.~Rubinstein, and K.~Aberman, ``Dreambooth: Fine tuning text-to-image diffusion models for subject-driven generation,'' in \emph{Proceedings of the IEEE/CVF Conference on Computer Vision and Pattern Recognition}, 2023, pp. 22\,500--22\,510.

\bibitem{micikevicius2017mixed}
P.~Micikevicius, S.~Narang, J.~Alben, G.~Diamos, E.~Elsen, D.~Garcia, B.~Ginsburg, M.~Houston, O.~Kuchaiev, G.~Venkatesh \emph{et~al.}, ``Mixed precision training,'' \emph{arXiv preprint arXiv:1710.03740}, 2017.

\bibitem{liu2024alora}
Z.~Liu, J.~Lyn, W.~Zhu, X.~Tian, and Y.~Graham, ``Alora: Allocating low-rank adaptation for fine-tuning large language models,'' \emph{arXiv preprint arXiv:2403.16187}, 2024.

\bibitem{farooq2024derm}
M.~A. Farooq, W.~Yao, M.~Schukat, M.~A. Little, and P.~Corcoran, ``Derm-t2im: Harnessing synthetic skin lesion data via stable diffusion models for enhanced skin disease classification using vit and cnn,'' \emph{arXiv preprint arXiv:2401.05159}, 2024.

\bibitem{abaid2024synthesizing}
A.~Abaid, M.~A. Farooq, N.~Hynes, P.~Corcoran, and I.~Ullah, ``Synthesizing cta image data for type-b aortic dissection using stable diffusion models,'' \emph{arXiv preprint arXiv:2402.06969}, 2024.

\bibitem{radford2021learning}
A.~Radford, J.~W. Kim, C.~Hallacy, A.~Ramesh, G.~Goh, S.~Agarwal, G.~Sastry, A.~Askell, P.~Mishkin, J.~Clark \emph{et~al.}, ``Learning transferable visual models from natural language supervision,'' in \emph{International conference on machine learning}.\hskip 1em plus 0.5em minus 0.4em\relax PMLR, 2021, pp. 8748--8763.

\bibitem{taited2023CLIPScore}
S.~Zhengwentai, ``{clip-score: CLIP Score for PyTorch},'' \url{https://github.com/taited/clip-score}, March 2023, version 0.1.1.

\bibitem{heusel2017gans}
M.~Heusel, H.~Ramsauer, T.~Unterthiner, B.~Nessler, and S.~Hochreiter, ``Gans trained by a two time-scale update rule converge to a local nash equilibrium,'' \emph{Advances in neural information processing systems}, vol.~30, 2017.

\bibitem{salimans2016improved}
T.~Salimans, I.~Goodfellow, W.~Zaremba, V.~Cheung, A.~Radford, and X.~Chen, ``Improved techniques for training gans,'' \emph{Advances in neural information processing systems}, vol.~29, 2016.

\bibitem{bińkowski2018demystifying}
\BIBentryALTinterwordspacing
M.~Bińkowski, D.~J. Sutherland, M.~Arbel, and A.~Gretton, ``Demystifying {MMD} {GAN}s,'' in \emph{International Conference on Learning Representations}, 2018. [Online]. Available: \url{https://openreview.net/forum?id=r1lUOzWCW}
\BIBentrySTDinterwordspacing

\bibitem{deng2018arcface}
J.~Deng, J.~Guo, X.~Niannan, and S.~Zafeiriou, ``Arcface: Additive angular margin loss for deep face recognition,'' in \emph{CVPR}, 2019.

\bibitem{van2008visualizing}
L.~Van~der Maaten and G.~Hinton, ``Visualizing data using t-sne.'' \emph{Journal of machine learning research}, vol.~9, no.~11, 2008.

\bibitem{10.3389/fpsyg.2014.01532}
\BIBentryALTinterwordspacing
V.~LoBue and C.~Thrasher, ``The child affective facial expression (cafe) set: validity and reliability from untrained adults,'' \emph{Frontiers in Psychology}, vol.~5, 2015. [Online]. Available: \url{https://www.frontiersin.org/journals/psychology/articles/10.3389/fpsyg.2014.01532}
\BIBentrySTDinterwordspacing

\bibitem{dosovitskiy2020image}
A.~Dosovitskiy, L.~Beyer, A.~Kolesnikov, D.~Weissenborn, X.~Zhai, T.~Unterthiner, M.~Dehghani, M.~Minderer, G.~Heigold, S.~Gelly \emph{et~al.}, ``An image is worth 16x16 words: Transformers for image recognition at scale,'' \emph{arXiv preprint arXiv:2010.11929}, 2020.

\bibitem{serengil2021lightface}
\BIBentryALTinterwordspacing
S.~I. Serengil and A.~Ozpinar, ``Hyperextended lightface: A facial attribute analysis framework,'' in \emph{2021 International Conference on Engineering and Emerging Technologies (ICEET)}.\hskip 1em plus 0.5em minus 0.4em\relax IEEE, 2021, pp. 1--4. [Online]. Available: \url{https://ieeexplore.ieee.org/document/9659697}
\BIBentrySTDinterwordspacing

\bibitem{zhifei2017cvpr}
Z.~Zhang, Y.~Song, and H.~Qi, ``Age progression/regression by conditional adversarial autoencoder,'' in \emph{IEEE Conference on Computer Vision and Pattern Recognition (CVPR)}.\hskip 1em plus 0.5em minus 0.4em\relax IEEE, 2017.

\end{thebibliography}

\vfill

\end{document}